\documentclass[conference]{IEEEtran}

\usepackage{booktabs}
\usepackage{multirow}
\usepackage{siunitx}
\usepackage{color}
\usepackage{epsf}
\usepackage{times}
\usepackage{epsfig}
\usepackage{epstopdf}
\usepackage{graphicx}
\usepackage{algorithm}
\usepackage{algorithmic}
\usepackage{amsmath}
\usepackage{amssymb}
\usepackage{amsxtra}
\usepackage{multirow}
\usepackage{mathtools}
\usepackage{subcaption}
\usepackage{multirow}
\usepackage{comment}
\usepackage[normalem]{ulem}
\usepackage{tabu}
\usepackage[dvipsnames]{xcolor}
\usepackage{multicol}
\usepackage{cite}
\usepackage{caption}
\usepackage{multirow}
\IEEEoverridecommandlockouts

\title{A Generative Graph Method to Solve the Travelling Salesman Problem}

\author{
\IEEEauthorblockN{\large Amal Nammouchi, Hakim Ghazzai, and Yehia Massoud}\\
\IEEEauthorblockA{\small School of Systems and Enterprises -- Stevens Institute of Technology, Hoboken, NJ, USA\\ Email: \{anammouc, hghazzai, ymassoud\}@stevens.edu\\
}
{\thanks {\hrule
\vspace{0.1cm} 
\indent This paper is accepted for publication in 63rd IEEE International Midwest
Symposium on Circuits and Systems (MWSCAS’20), Springfield, MA, USA, Aug. 2020.

© 2020 IEEE. Personal use of this material is permitted. Permission from
IEEE must be obtained for all other uses, in any current or future media,
including reprinting/republishing this material for advertising or promotional
purposes, creating new collective works, for resale or redistribution to servers
or lists, or reuse of any copyrighted component of this work in other works.

}}\vspace{-0.0cm}}
\begin{document}
\maketitle
\thispagestyle{empty}

\begin{abstract}
\boldmath{
The Travelling Salesman Problem (TSP) is a challenging graph task in combinatorial optimization that requires reasoning  about both local node neighborhoods and  global  graph  structure. In this paper, we propose to use the novel Graph Learning Network (GLN), a generative approach, to approximately solve the TSP. GLN model learns directly the pattern of TSP instances as training dataset, encodes the graph properties, and merge the different node embeddings to output node-to-node an optimal tour directly or via graph search technique that validates the final tour. The preliminary results of the proposed novel approach proves its applicability to this challenging problem providing a low optimally gap with significant computation saving compared to the optimal solution.
}
\end{abstract}\vspace{0.1cm}


\section{Introduction}
\label{Sec1a}

The Travelling Salesman Problem (TSP) is one of the most intensively studied problems and widely used as a benchmark in operations research, computational mathematics, and computer science. Precisely, TSP states a salesman and $n$ cities defined as nodes on an undirected graph where the objective is to find the optimal tour with the minimum distance length visiting every node once and only once and returning to the starting node. Solving the TSP seems fairly simple, but as it is a known NP-hard problem, i.e., cannot be solved in polynomial time, it is rather hard to solve. Given this property, the problem is usually solved by methods that lead to suboptimal approximate solutions in reasonable computational time. Typically, designed approaches to solve these kinds of combinatorial optimization problems can be divided into two groups: exact methods and heuristics. The exact methods guarantee ﬁnding optimal solutions if allowed to completely perform their search. These algorithms have high execution cost making them unsuitable for large instances. These optimal solutions, e.g., the branch-and-bound algorithm, are usually implemented in off the shelf software such as the Concorde solver~\cite{01227} and Gurobi. As for heuristics, they designed to trade
off optimality for computational cost. Such solutions include nearest insertion, nearest neighbor, and farthest insertion, etc.

Despite being in the NP-hard class problems, the TSP problem never ceased to attract the research community thanks to its importance in many real-life applications mainly in  transportation  and  logistics.  In  fact,  in  many fields  of  operation,  it  is  important  to  find  a  shortest  path through  a  collection  of  points.  The  most  common  nowadays examples  specifically  in  the  industry  consist  in  the  supply chain,  packages  picking  and  stocking  in  huge  warehouses where  a  solution  as  the  one  of  TSP  might  optimize the delivery time  as well as the shortest  route  so as to  save fuel and labor cost. As a result, a growing interest appeared over years among the Artificial Intelligence community to solve optimization problems using Deep Learning (DL) methods. One of the ﬁrst artiﬁcial neural networks designed to solve the TSP is the model known as Hopfield networks ~\cite{article}. An overview of some other first trials can be found in ~\cite{6419953}. These historical studies though are still dissatisfying compared to the heuristic algorithms in terms of speed and optimality. Recently, Graph Neural Network (GNN)  has attracted a lot of attention~\cite{4700287,9046288}, and the existing research has shown the efficacy of graph learning methods to leverage the graph structure and solve many problems. Consequently, since TSP is naturally modeled in graphs, using GNN to solve the computational hardness of optimization problems certainly shows promising results.

Recent work using deep learning methods on graphs to solve TSP can be divided into auto-regressive and non-autoregressive approaches. An auto-regressive model generates node-to-node graphs like the Graph Attention Network used in~\cite{333} which is attention-based decoder trained with reinforcement learning to autoregressively build TSP solutions. The work in~\cite{kool2018attention} uses a more powerful decoder and trained the model using REINFORCE~\cite{111} with a greedy rollout baseline to achieve state-of-the-art results among learning-based approaches for TSP. In~\cite{nowak2017revised}, a supervised approach using GNN, the output is a tour presented as an adjacency matrix, which is converted into a valid tour using beam search. However, it performs poorly for very small problem instances. 

In this paper, we introduce a generative graph learning approach for approximately solving
the TSP and validating the final tour using the Graph Learning Network (GLN). GLN is a graph learning technique introduced in~\cite{Saire2019} which the authors evaluate as an edge classifier and simulate its performances as a graph generator to predict the graph structure based on learned-patterns. They tested it on synthetic data to predict the structure of geometric and 3D figures dataset. In our case, the TSP solution pattern is learned by training the model with a generated family of graphs representing the TSP instances. Afterwards, we predict the TSP graph from the local and global node embedding in the graph through a recurrent set of operations. This approach presents a significant amelioration in speed solving time compared to optimal methods, as well as the efficiency of the solution with less training data and smaller graphs. In fact, the input to initiate the training consists in the node feature vector and an aleatory adjacency matrix, while most of the other learning solutions use full graphs as an input. Finally, although we return to the search technique in order to validate the final tour, our approach directly output the optimal tour adjacency matrix in many cases.

\begin{figure}[t!]
        \centerline{\includegraphics[width=9cm]{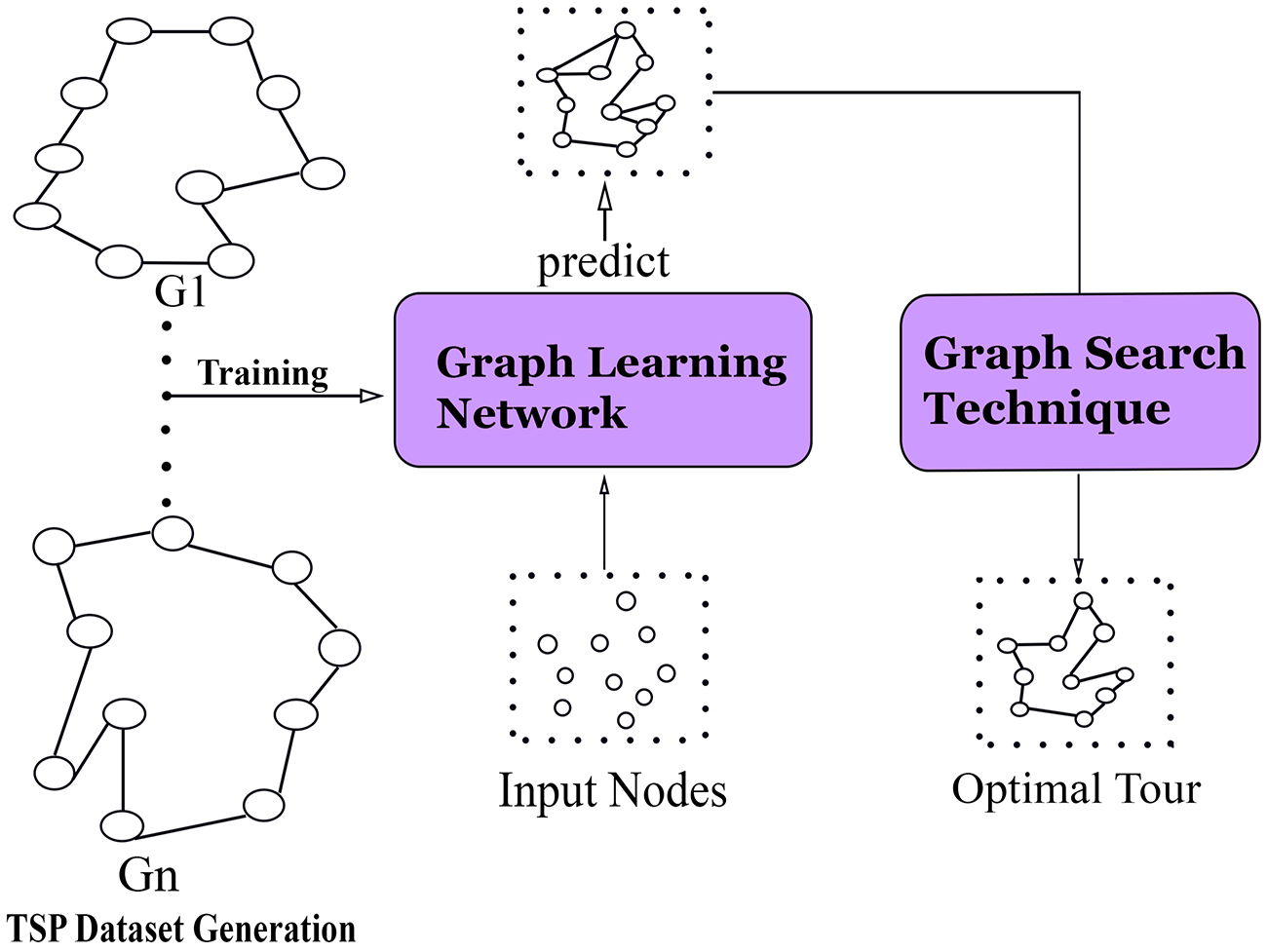}}
    \caption{Proposed GLN-TSP approach for solving the Travelling Salesman Problem.}
\vspace{-0.3cm}
\label{fig:framework}
\end{figure}
\section{Graph-TSP Framework}\label{sec3}


 Figure 1 presents an overview of our proposed approach. Our model takes a set of vertices and their feature vector as an input and predicts the optimal graph structure. In other words, it predicts the set of edges between those vertices, i.e., the adjacency matrix. We then convert the output graph into a valid tour via a graph search technique. Unlike previous work \cite{DBLP:journals/corr/abs-1906-01227} that trains their models in a supervised manner using pairs of problem instances and optimal solutions, we directly feed the generated TSP graphs to the model which learns their patterns in order to propose the best expected node features for the target nodes. The representations that we learn through the graph convolutions can encode properties of the graph structure, learn the best structure, and thus approximately predict the optimal graph tour.

\subsection{TSP Data Set Generation}
A particular case in TSP is the Euclidean TSP, which is the main interest of this paper. In this version, each node is assigned coordinates in a plane, and the cost on an edge connecting two nodes is the Euclidean distance between them. In this work, we generate the TSP data set using Concorde Solver, which outputs a family of graphs that represents the optimal solutions of different cases. The node features of each graph are the coordinates of the nodes that we randomly generate and then normalize to be $\in [0, 1]$.

We generate 50000 graphs of different sizes: 10, 20, 30, and 50 node instances that we split into training, validation, and testing.

\subsection{GLN Model}
Given a set of vertices $\mathcal V= \{v_{i}\}$, e.g., it can represent a set of cities, where $v_{i}$ is the feature vector of each node representing the node location in the 2D space in our case. We train the model to leverage the structure of the graph and learn the possible set of edges $\mathcal E=\{e_{ij}\}$ that maximizes the relations between the vertices, i.e., minimizes the Euclidean distances separating them, and hence, the final tour length. Each graph will be used to learn the parameters of the GLN model that minimizes a loss function that will presented in details later.

Based on Graph Convolutional Networks introduced in~\cite{kipf2016semisupervised}, the GLN model leverages the structure of the graph by extracting both local and global representations, denoted by $ H_{local}^{(l)}$ and $H_{global}^{(l)}$ at each step $l$, and combining them to predict the graph structure, i.e., the adjacency matrix, denoted by $A^{(l+1)}$ in a recursive way. Algorithm 1 presents the pseudo-code of the adjacency matrix prediction procedure using the following equations:
\begin{align}
  &  H_{local}^{(l)} = ReLu\left({A^{(l)}} H_{int}^{(l)} U^{(l)}\right),\\
  &\text{where } H_{int}^{(l)} = \sum_{i=1}^{k} ReLu\left({A^{(l)}} H_{i}^{(l)} W_{i}^{(l)}\right),\\
&     H_{global}^{(l)} = ReLu\left( H_{local}^{(l)} Z^{(l)}\right),\\
&   \text{and } A^{(l+1)} = ReLu\left(M^{(l)} H_{local}^{(l)} Q^{(l)} {\left(H_{global}^{(l)}\right)}^\top  {\left(M^{(l)}\right)}^\top\right),
\end{align}
where $W_{i}^{(l)}$, $U^{(l)}$, $Z^{(l)}$, $M^{(l)}$, and $Q^{(l)}$ are learnable matrices at each step $l$. The notation $ReLu$ denotes the rectified linear unit. Finally, $\lambda_{l}$, $\gamma_{l}$, and $\alpha_{l}$ used in Algorithm 1 are embedding functions introduced in~\cite{Saire2019}.\\
\begin{algorithm}[t]
\caption{Node Embeddings Generation and Adjancecy Matrix Prediction (Recurrent Block)}
\begin{algorithmic}[1]
\STATE \textbf{Input: }\texttt{Feature Vector $X_v \in \mathbb{R}_{}^{n \times d_{l}}, \forall v \in V$ , d is the feature dimension, k the number of kernels }
\STATE \textbf{Output:}\texttt{ Vector Representation  $H_{}^{(L)}$,Adjacency Matrix  $A_{}^{(L)}$}
\STATE $H_{}^{(0)} \gets X$
\STATE $A_{}^{(0)} \gets I_{n}$
\STATE \textbf{for} $l= 1 ... L$ \textbf{do}
\STATE \hspace{0.4cm}\textbf{for} $i= 1 ... k$ \textbf{do}
\STATE \hspace{0.4cm}\hspace{0.4cm}\textbf{extract features $H_{i}^{(l)}$}
\STATE \hspace{0.4cm} $H_{int}^{(l)} \gets \textbf{combine} (H_{1}^{(l)} ... H_{k}^{(l)})$
\STATE \hspace{0.4cm} $H_{local}^{(l)} \gets \lambda_{l} (H_{int}^{(l)}, A_{}^{(l)})$
\STATE \hspace{0.4cm} $H_{global}^{(l)} \gets \gamma_{l} H_{local}^{(l)}$
\STATE \hspace{0.4cm} $H_{}^{(l+1)} \gets  H_{local}^{(l)}$
\STATE \hspace{0.4cm} $A_{}^{(l+1)} \gets \alpha_{l} (H_{local}^{(l)},H_{global}^{(l)})$
\end{algorithmic}
\end{algorithm} 
$\bullet$\textbf{ Initialization: }Given the feature vector as the only input, the model needs some structure to be initiated with for training. In this work, $A_{}^{(0)}$ is an aleatory matrix declared with $p$ edges ($p$ values of $1$ and $1-p$ values of $0$). However, other structures can be used as well such as the identity matrix $I_{n}$.\\
$\bullet$\textbf{ Loss Function: } Aware of the high sparsity of the TSP problem and as we deal with it as edge classification where the class of edges, i.e., class of $1$ is less than the class of non-edges, i.e., class of $0$, we tend to use a weighted loss function. Hence, we aim to minimize the following loss function:
\begin{equation}
   L_{total} =\psi_{1} L_{hed} + \psi_{2}L_{IoU},
\end{equation}
where $L_{hed}$ is the Hed-loss function introduced in~\cite{Xie_2015_ICCV} which is a class-balanced cross-entropy function that penalizes each class prediction while $L_{IoU}$ is the intersection-over-union loss~\cite{Milletari2016VNetFC} such as  we treat the edges on the adjacency matrices as regions on an image and we compare the whole structure of the predicted graph with its ground truth.
Finally, $\psi_{1} $ and $\psi_{2} $ are hyper-parameters that deﬁne the contribution of each loss to the learning process.

\subsection{Graph Search Technique}
The model outputs an adjacency matrix with a deviation ranging from $0\%$ to $3\%$ from the ground truth matrix.
In fact, the model is able to directly produce the optimal graph. However, it may yield few extra edges in some cases. That is when we need to valid the final tour via the search technique. Many search strategies be can employed in this case, yet we opt for greedy search. In general, greedy algorithms choose the local optimal solution to provide a fast
approximation of the global optimal solution. Starting from the ﬁrst node (the starting point of each tour), we greedily select the next node from its neighbors based on the euclidean distance. The search
terminates when all nodes have been visited. We mask out nodes that have previously been visited in
order to construct valid solutions.

\section{Results and Discussions}\label{sec4}
In this section, we present and discuss the obtained results. We describe the training phase by giving an insight about the different used hyper-parameters and data-set. We also describe the evaluation process of the model which is evaluated as an edge classifier and simulated as a graph generator. Finally, we test the model performances and compare it to existing studies.

\subsection{Training phase}
We focus on solving TSP problems. Nevertheless, our method can be easily applied to other routing problems. We train the model on 50000 graphs of different instances sizes: $n \in \{10, 20, 30, 50\}$. The node features correspond to randomly generated 2D coordinates (the solution can be applied to 3D space too).
For each problem size, we use the same hyper-parameters. We run the model with a recurrent block using hidden values of 32 for three levels of depth (layers). We use the Adam optimizer and a fixed learning rate of $0.001$ while we use a batch size equal to $50$. We return to an early stopping strategy in order to avoid the model overfitting.

Figure 2 shows the evolution of the total loss function given in (5) during the training for different sizes of TSP instances ($n=20$ and $n=30$). At the beginning, we notice a cold start but then the model rapidly converges. The convergence speed depends on the size of the graph.

\begin{figure}[t!]
        \centerline{\includegraphics[width=9cm, height=4.5cm]{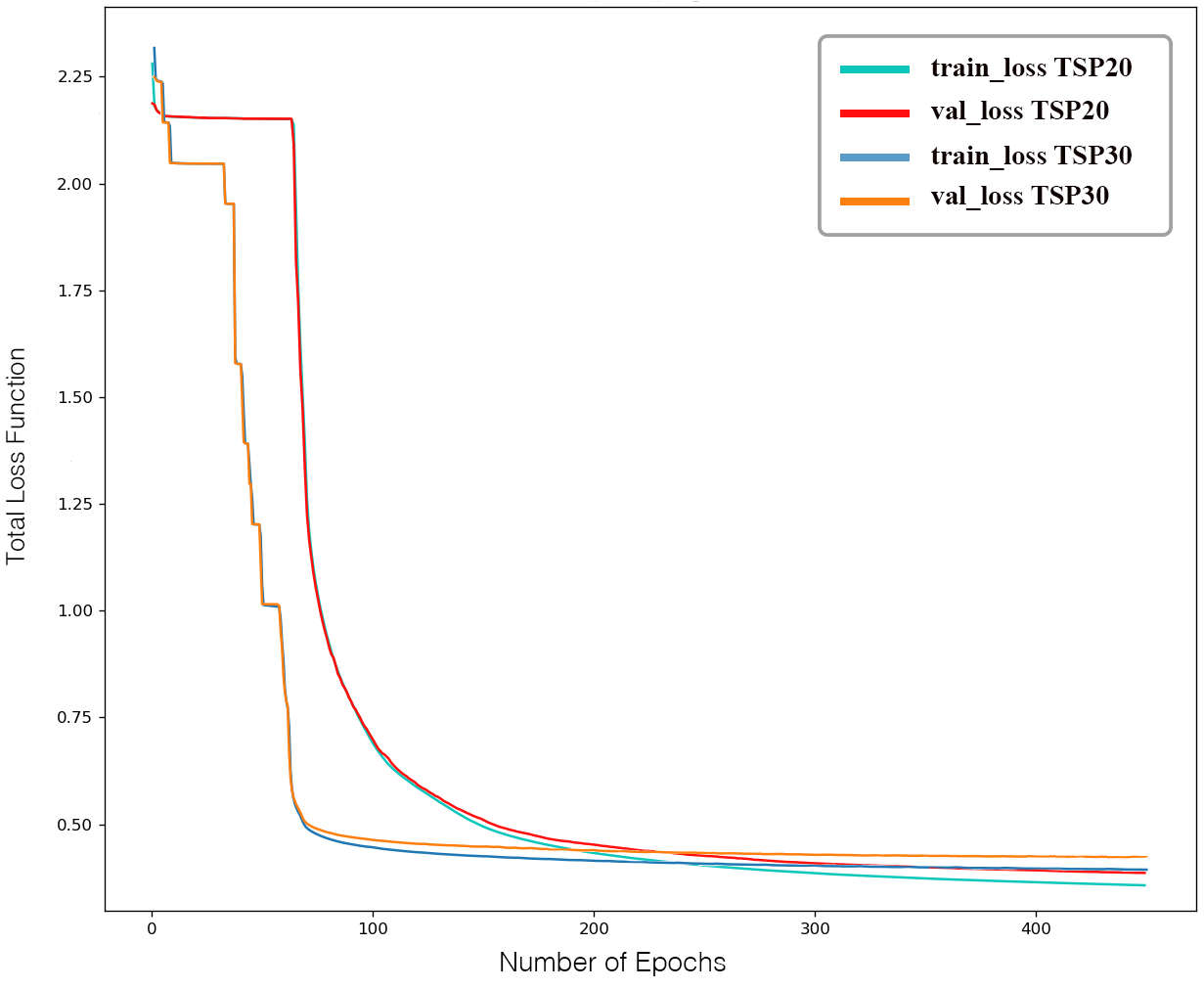}}
    \caption{The total training and validation loss function for variant size instances.}
\vspace{-0.3cm}
\label{fig:framework}
\end{figure}

\subsection{Performance Comparisons}
We test the model on $10000$ instances. The model is evaluated as an edge classifier and considers as evaluation metric the $F1$-score, which is expressed as follows:
\begin{align}
&F1\textsc{-}Score = \frac{2(Precision\times Recall)}{Precision+Recall},\\
& \text{where }Precision = \frac{TP}{TP+FP} \text{ and } Recall = \frac{ TP +FN}.\notag
\end{align}
We intend to maximize these metrics in order to predict the best structure, i.e, the optimal tour. In many cases mainly for small scale graph, the model is able to directly produce the optimal graph node-by-node with $0\%$ deviation from the ground truth.

\begin{table*}[t]
\begin{center}
\caption{\label{tab1} Performance on TSP testing set}
 \begin{tabular}{||p{6mm}||p{8mm}||p{7mm}||p{7mm}|p{11mm}|p{7mm}||p{8mm}|p{7mm}|p{12mm}|p{12mm}||p{7mm}|p{7mm}|p{7mm}|p{7mm}||}
 \hline
 Model&K-NN Heuristics&MLP &GCN &GraphSage & GIN & GAT &MoNet &GatedGCN &GatedGCN-E* &GLN-TSP10 &GLN-TSP20 &GLN-TSP30 &GLN-TSP50 \\
 \hline
  F1-Score&0.693&0.548 &0.627 &0.663 & 0.657 & 0.669 &0.637 & 0.794 &0.802 &0.872 &0.827  & 0.763  &0.711 \\
   \hline
 
 \end{tabular}
\end{center}  \vspace{-0.2cm}
\end{table*}

\begin{table*}[t]
  \begin{center}
  \caption{\label{tab2} Performance of the GLN-TSP vs. non-learned baselines and state-of-the-art methods for various TSP instance sizes}
  \begin{tabular}{||p{40mm}| p{30mm}p{30mm}| p{30mm}p{30mm}||}
    \toprule
    \multirow{3}{*}{Model} &
      \multicolumn{2}{c|}{TSP20 } &
      \multicolumn{2}{c||}{TSP50 } \\
      & {Tour Len} & {Opt.Gap}  & {Tour Len} & {Opt.Gap}  \\
      \midrule
    Concorde Solver &  3.84 & 0.00\%  & 5.70 & 0.00\%   \\
    LKH3 Solver &  3.84& 0.00\% & 5.70& 0.00\%  \\
    Gurobi Solver & 3.84 &0.00\% &5.70& 0.00\%  \\
    \midrule
    Nearest Insertion (H, G)  & 4.33 &12.91\% & 6.78& 19.03\%  \\
    Random Insertion (H, G) & 4.00& 4.36\%  &6.13& 7.65\%\\
    Farthest Insertion (H, G)  &  3.93& 2.36\% & 6.01& 5.53\%  \\
    Nearest Neighbor (H, G) &  4.50& 17.23\% & 7.00& 22.94\%  \\
    PtrNet (SL, G) &  3.88& 1.15\%& 7.66& 34.48\% \\
    PtrNet (RL, G) &   3.89 &1.42\% &5.95& 4.46\% \\
    
    \midrule
    GAT (RL, S) &  3.84 & 0.08\%&  5.73& 0.52\%  \\
    GAT (RL, G)  &3.85 &0.34\%  &5.80& 1.76\%\\

    \midrule
   \textbf{GLN-TSP} & 3.85 & 0.34\%  & 5.85 & 2.78\% \\

    \bottomrule
  \end{tabular}
  \end{center}
  \vspace{-0.2cm}
\end{table*}
Table~\ref{tab1} shows the obtained $F1$-score of the TSP testing set. The performances of the different edge classifiers are taken from the GNN Benchmarking~\cite{dwivedi2020benchmarking} for the goal of comparison. As we deal with sparse graph instances while other models consider a fixed input full graph which makes it an exhaustive process especially for large-scale instances, we notice that, in our case, the $F1$-score decreases as the problem size increases, however, still well-performing.

In order to evaluate the prediction performances and compare it with previous work, we return to the following evaluation procedure used previously in~\cite{kool2018attention}:\\
$\bullet$\textbf{ Predicted tour length: }the average predicted TSP tour length, denoted by $Tour\text{-}Len$, over 10,000 test instances expressed as follows:
\begin{equation}
    Tour\text{-}Len = 
    \frac{1}{m}\sum_{i=1}^{m}{l_{tsp}(i)},
\end{equation}
where $m=10000$ and $l_{tsp}$ is the tour length of graph instance~$i$.\\
$\bullet$\textbf{ Optimality gap: }the average percentage ratio of the predicted tour length, $\hat{l}_{tsp}$ relative to the optimal tour of the solver $l_{tsp}$. It is denoted by $ Opt\text{-}Gap$ and expressed as follows:

\begin{equation}
    Opt\text{-}Gap = 
    \frac{1}{m}\sum_{i=1}^{m}\left(\frac{{l_{tsp}(i)}}{{\hat{l}_{tsp}(i)}}-1\right).
\end{equation}
Table~\ref{tab2} shows the performance of the proposed model (GLN-TSP) in comparison with deep learning techniques, exact solvers and heuristic methods where H, G, SL, RL, and D denote Heuristic, Greedy, Supervised Learning, Reinforcement Learning and Sampling, respectively. The proposed model outperforms simple heuristics and presents a low optimal gap mainly thanks to the model's ability to generate optimal tours directly in most of the cases without deviation from the ground truth. Our main goal is to propose a novel approach highly competitive to the existing heuristics and deep learning techniques to approximately solve combinatorial problems in a generative way. In fact, the preliminary results of the GLN-TSP proves the potential of the model to learn the graph aspect and solve the TSP with few training data instances while other studies like \cite{DBLP:journals/corr/abs-1906-01227} used one Million pairs of full graphs and solution instances to train their model, we only use 50000 instances for the training and yet achieve close results.

\section{Conclusion and Future Work}\label{sec5}
In this paper, we introduced a simple yet effective approach for approximately solving the 2D Euclidean Travelling Salesman Problem using Graph Learning Network and greedy graph search. TSP’s multi-scale nature makes it a challenging graph task which requires reasoning about both local node neighborhoods and global graph structure which we successfully tackled using the GLN model. 
In future work, it is important to incorporate methods to deal with sparse graphs in order to generalize our method to large-scale problem instances, as well as ameliorate the model performance to become able to directly output the optimal tour for all the cases without returning to graph search technique to valid the final tour and thus, unleash the powerful side of our method. Finally, applying the developed approach on TSP generalization and other routing problems will be considered in the future extension of this work.


\bibliographystyle{IEEEtran}
\bibliography{References}

\begin{thebibliography}{10}
\providecommand{\url}[1]{#1}
\csname url@samestyle\endcsname
\providecommand{\newblock}{\relax}
\providecommand{\bibinfo}[2]{#2}
\providecommand{\BIBentrySTDinterwordspacing}{\spaceskip=0pt\relax}
\providecommand{\BIBentryALTinterwordstretchfactor}{4}
\providecommand{\BIBentryALTinterwordspacing}{\spaceskip=\fontdimen2\font plus
\BIBentryALTinterwordstretchfactor\fontdimen3\font minus
  \fontdimen4\font\relax}
\providecommand{\BIBforeignlanguage}[2]{{%
\expandafter\ifx\csname l@#1\endcsname\relax
\typeout{** WARNING: IEEEtran.bst: No hyphenation pattern has been}%
\typeout{** loaded for the language `#1'. Using the pattern for}%
\typeout{** the default language instead.}%
\else
\language=\csname l@#1\endcsname
\fi
#2}}
\providecommand{\BIBdecl}{\relax}
\BIBdecl

\bibitem{01227}
V.~C. David L~Applegate, Robert E~Bixby and W.~J. Cook, ``The traveling
  salesman problem: a computational study,'' \emph{Princeton university press},
  2006.

\bibitem{article}
J.~Hopfield and D.~Tank, ``Neural computation of decisions in optimization
  problems,'' \emph{Biological cybernetics}, Feb 1985.

\bibitem{6419953}
B.~F.~J. {La Maire} and V.~M. {Mladenov}, ``Comparison of neural networks for
  solving the travelling salesman problem,'' in \emph{11th Symposium on Neural
  Network Applications in Electrical Engineering}, Belgrade, Serbia, 2012.

\bibitem{4700287}
F.~{Scarselli}, M.~{Gori}, A.~C. {Tsoi}, M.~{Hagenbuchner}, and
  G.~{Monfardini}, ``The graph neural network model,'' \emph{IEEE Transactions
  on Neural Networks}, Jan. 2009.

\bibitem{9046288}
Z.~{Wu}, S.~{Pan}, F.~{Chen}, G.~{Long}, C.~{Zhang}, and P.~S. {Yu}, ``A
  comprehensive survey on graph neural networks,'' \emph{to appear in IEEE
  Transactions on Neural Networks and Learning Systems}, 2020.

\bibitem{333}
M.~Deudon, P.~Cournut, A.~Lacoste, Y.~Adulyasak, and L.-M. Rousseau, ``Learning
  heuristics for the {TSP} by policy gradient,'' in \emph{Integration of
  Constraint Programming, Artificial Intelligence, and Operations Research},
  W.-J. van Hoeve, Ed.\hskip 1em plus 0.5em minus 0.4em\relax Cham: Springer
  International Publishing, 2018.

\bibitem{kool2018attention}
W.~Kool, H.~van Hoof, and M.~Welling, ``Attention, learn to solve routing
  problems!'' in \emph{International Conference on Learning Representations},
  New Orleans, LA, USA, May 2019.

\bibitem{111}
R.~J. Williams, ``Simple statistical gradient-following algorithms for
  connectionist reinforcement learning,'' \emph{Machine Learning}, vol.~8, pp.
  229--256, May 1992.

\bibitem{nowak2017revised}
A.~{Nowak}, S.~{Villar}, A.~S. {Bandeira}, and J.~{Bruna}, ``Revised note on
  learning quadratic assignment with graph neural networks,'' in \emph{IEEE
  Data Science Workshop (DSW'18)}, Lausanne, Switzerland, June 2018.

\bibitem{Saire2019}
D.~Saire and A.~Ram\'irez~Rivera, ``Graph learning network: A structure
  learning algorithm,'' June, 2019.

\bibitem{DBLP:journals/corr/abs-1906-01227}
C.~K. Joshi, T.~Laurent, and X.~Bresson, ``An efficient graph convolutional
  network technique for the travelling salesman problem,'' June, 2019.

\bibitem{kipf2016semisupervised}
T.~N. Kipf and M.~Welling, ``Semi-supervised classification with graph
  convolutional networks,'' in \emph{International Conference on Learning
  Representations (ICLR'17)}, Toulon, France Apr. 2017.

\bibitem{Xie_2015_ICCV}
S.~Xie and Z.~Tu, ``Holistically-nested edge detection,'' in \emph{IEEE
  International Conference on Computer Vision (ICCV'15)}, Las Condes, Chile
  Dec. 2015.

\bibitem{Milletari2016VNetFC}
F.~Milletari, N.~Navab, and S.-A. Ahmadi, ``{V-Net}: Fully convolutional neural
  networks for volumetric medical image segmentation,'' \emph{Fourth
  International Conference on 3D Vision (3DV'16)}, Stanford Univeristy, CA,
  USA, Oct. 2016.

\bibitem{dwivedi2020benchmarking}
V.~P. Dwivedi, C.~K. Joshi, T.~Laurent, Y.~Bengio, and X.~Bresson,
  ``Benchmarking graph neural networks,'' Mar. 2020.

\end{thebibliography}
\end{document}